\documentclass[sn-apa]{sn-jnl}

\usepackage{graphicx}%
\usepackage{multirow}%
\usepackage{amsmath,amssymb,amsfonts}%
\usepackage{amsthm}%
\usepackage{mathrsfs}%
\usepackage[title]{appendix}%
\usepackage{xcolor}%
\usepackage{textcomp}%
\usepackage{manyfoot}%
\usepackage{booktabs}%
\usepackage{algorithm}%
\usepackage{algorithmicx}%
\usepackage{algpseudocode}%
\usepackage{listings}%
\usepackage{appendix}
\usepackage{array}

\theoremstyle{thmstyleone}%
\theoremstyle{thmstyletwo}%

\theoremstyle{thmstylethree}%

\begin{document}

\title[Article Title]{Evaluating Soccer Match Prediction Models: A Deep Learning Approach and Feature Optimization for Gradient-Boosted Trees
}


\author[1]{\fnm{Calvin} \sur{Yeung}}\email{yeung.chikwong@g.sp.m.is.nagoya-u.ac.jp}

\author[1]{\fnm{Rory} \sur{Bunker}}\email{rory.bunker@g.sp.m.is.nagoya-u.ac.jp}

\author[1]{\fnm{Rikuhei} \sur{Umemoto}}\email{umemoto.rikuhei@g.sp.m.is.nagoya-u.ac.jp}

\author*[1,2,3]{\fnm{Keisuke} \sur{Fujii}}\email{fujii@i.nagoya-u.ac.jp}

\affil[1]{\orgdiv{Graduate School of Informatics}, \orgname{Nagoya University}, \orgaddress{\city{Nagoya}, \country{Japan}}}

\affil[2]{\orgdiv{Center for Advanced Intelligence Project}, \orgname{RIKEN}, \orgaddress{\city{Osaka}, \country{Japan}}}

\affil[3]{\orgdiv{PRESTO}, \orgname{Japan Science and Technology Agency}, \orgaddress{\city{Saitama}, \country{Japan}}}


\abstract{
Machine learning models have become increasingly popular for predicting the results of soccer matches, however, the lack of publicly-available benchmark datasets has made model evaluation challenging. The 2023 Soccer Prediction Challenge required the prediction of match results first in terms of the exact goals scored by each team, and second, in terms of the probabilities for a win, draw, and loss. The original training set of matches and features, which was provided for the competition, was augmented with additional matches that were played between 4 April and 13 April 2023, representing the period after which the training set ended, but prior to the first matches that were to be predicted (upon which the performance was evaluated). A CatBoost model was employed using pi-ratings as the features, which were initially identified as the optimal choice for calculating the win/draw/loss probabilities. Notably, deep learning models have frequently been disregarded in this particular task. Therefore, in this study, we aimed to assess the performance of a deep learning model and determine the optimal feature set for a gradient-boosted tree model. The model was trained using the most recent five years of data, and three training and validation sets were used in a hyperparameter grid search. The results from the validation sets show that our model had strong performance and stability compared to previously published models from the 2017 Soccer Prediction Challenge for win/draw/loss prediction. 
}




\maketitle

\section{Introduction}\label{sec1}
Soccer, also known as ``Association Football'' or ''Football'', is widely recognized as the most popular sport worldwide in terms of both spectatorship and player numbers. 
As a generally low-scoring sport, especially at the professional level, small goal margins, competitive leagues, and draws being a common outcome make predicting soccer match results a challenging task \citep{bunker2022application,yeung2023framework}, especially when only goals are available in the dataset \citep{berrar2019incorporating}.
The inherent unpredictability of outcomes is, of course, one of the primary reasons soccer attracts such a large number of fans.
Despite its challenging nature, given the popularity of the sport, there is a wide range of stakeholders who are interested in the prediction of soccer match results including fans, bookmaking companies, bettors, media, as well as coaches and performance analysts.

Models for match result prediction were historically proposed from disciplines such as statistics \citep{maher1982modelling,dixon1997modelling}, operations research, and mathematics.
However, with the advent of machine learning (ML), which is a sub-discipline of computer science that uses many techniques from statistics, over the past two decades, ML models have become a popular approach to predict match outcomes in soccer.
The lack of publicly-available benchmark datasets has, however, meant that it has been challenging for researchers to evaluate their results against other studies.
Match features used in models, which are derived from events that occur within matches, are often contained in vendor-specific event data streams that are generally only available to professional teams \citep{decroos2019actions}.
The Open International Soccer Database \citep{dubitzky2019open}, despite not containing such match features, has enabled researchers to compare their models in a like-for-like manner on a large number of soccer matches (over 216,000 matches across 52 leagues).
The 2017 Soccer Prediction Challenge \citep{berrar2019guest} was held, with participants using the Open International Soccer database to predict 206 unplayed matches.
Some of the top-ranked participants in the 2017 Soccer Prediction Challenge used gradient-boosted tree models and/or rating features \citep{berrar2019incorporating,hubavcek2019learning,constantinou2019dolores}, which suggested that condensing a wide range of historical match information into ratings was of benefit, as was using the accuracy-enhancing benefits of boosting.
Subsequently, other studies \citep{razali2022football,razali2022deep,robberechts2019forecasting} have used the Open International Soccer Database, in some cases improving upon the 2017 challenge results \citep{razali2022football}.
%

Deep learning has, over the past few years, gained in popularity for the prediction of match results in soccer, given its success in many domains including computer vision, trajectory analysis, and natural language processing.
In soccer, deep learning has been helpful in predicting the locations and types of subsequent events \citep{simpson2022seq2event,yeung2023transformer}, the outcomes of shots \citep{yeung2023strategic}, and in using match video to detect and track players and/or the ball, to detect events, and to analyze matches \citep{akan2023use}.
Two types of models have thus emerged as potential state-of-the-art models for sports match result prediction: deep learning and boosted decision tree models.

A subsequent soccer prediction challenge competition was held in 2023,\footnote{\url{https://sites.google.com/view/2023soccerpredictionchallenge}} using a similar dataset.
However, unlike the 2017 Soccer Prediction Challenge, the 2023 competition required two tasks, the first of which, ``exact score prediction,'' involved predicting match results in terms of exact goals scored (for each team), and the second, ``probability prediction,'' involved prediction in terms of the probabilities for a win, draw, and loss.

In the current study, initially, consistent with \cite{razali2022football}, a CatBoost model \citep{prokhorenkova2018catboost}  with pi-ratings \citep{constantinou2013determining} as the model features was found to be the best-performing model for win/draw/loss probability prediction.
However, to explore the potential of deep learning models for match result prediction in soccer, we then developed a deep learning-based model for win/draw/loss probability prediction that utilizes a combination of cutting-edge techniques.
Specifically, the proposed method incorporates modules from the TimesNet time series model \citep{wu2022timesnet}, Transformer, a neural language processing model \citep{vaswani2017attention}, and a neural network. 
The model was trained using the most recent five years of data available in the challenge dataset.
Furthermore, to compute the prediction set features, we augmented the matches and features in the training set provided as part of the competition with additional matches that were played between 4 April and 13 April 2023.
This nine-day period represented an ``in-between period'': the dates after the end of the training set but prior to the first match date in the evaluation set to be predicted.
A grid search was used to select optimal hyperparameters by training the models with these various hyperparameters on three training sets and evaluating their performance on three validation sets.
The results from the validation sets show that our model outperformed all previously published models from the 2017 Soccer Prediction Challenge for win/draw/loss probability prediction.
%

The main contributions of this study are as follows. 
First, we investigate whether our developed deep learning models are superior to existing state-of-the-art models. 
Second, we examine how deep learning models can be applied for match result prediction in terms of learning the time series nature of the data and engineering additional features.
Finally, the proposed deep learning-based models are compared with existing models and real-world data are used to evaluate the effectiveness of the approach.

The remainder of this paper is organized as follows.
In Section \ref{sec:methods}, we detail the model used in this study for soccer match result prediction.
Then, the experimental results are presented and discussed in Section \ref{sec:results}.
Following this, we discuss research related to the current study in Section \ref{sec:related}, including the existing literature related to the 2017 Soccer Prediction Challenge, as well as studies subsequent to the competition that also used the Open International Soccer Database.
Finally, the paper is concluded in Section \ref{sec:conclusion}.

\section{Methods}
\label{sec:methods}
In this section, we first give a more detailed introduction to the 2023 Soccer Prediction Challenge in Section \ref{sec:2023_spc}. Then, we introduce the deep learning and boosted decision tree methods in Sections \ref{sec:approach_dl} and \ref{sec:approach_tree}, respectively. Lastly, in Section \ref{sec:dataset_split}, we explain the training and validation methods for the models.

\subsection{2023 Soccer Prediction Challenge}
\label{sec:2023_spc}
As mentioned in the introduction, the 2023 Soccer Prediction Challenge made available to participants a dataset similar to that of the 2017 competition to train their models. This training dataset comprised match results from 51 soccer leagues from 2001 to April 4, 2023, encompassing a total of over 300,000 matches. The dataset of matches to be predicted included 736 matches from 44 leagues, spanning the period from April 14 to April 26, 2023.
Within these datasets, nine distinct features were provided: the season, league, date of the match, names of the home and away teams, the goals scored by the home and away teams, the difference in goals scored between the home and away teams, and the outcome of the match(win/draw/loss).
Participating contestants were granted the option to leverage supplementary publicly available data. As mentioned in the introduction, the 2023 Soccer Prediction Challenge necessitated the completion of two distinct tasks: exact score prediction and probability prediction. 

\textbf{Task 1: Exact Scores Prediction.}
Predicting match results based on the exact goals scored by the home and away teams was a task introduced in the 2023 competition, which was not required in the 2017 competition.
The evaluation metric used for Task 1 was the Root Mean Squared Error (RMSE), which is defined as follows:
\begin{equation}
\label{eq1}
    \text{RMSE} = \sqrt{\frac{1}{N} \sum_{i=1}^{N}(y_i - \hat{y}_i)^2}
\end{equation}

\noindent where \(N\) denotes the total number of data points, \(y_i\) denotes the actual observed values, and \(\hat{y}_i\) represents the values predicted by the model.
Models with a lower RMSE are preferred.
%
%

\textbf{Task 2: Probabilities Prediction.}
The prediction of match results based on win, draw, and loss probabilities, which was the sole task in the 2017 Soccer Prediction Challenge, was the second task in the 2023 competition.
The evaluation metric used for Task 2 models was the Ranked Probability Score (RPS) \citep{epstein1969scoring,constantinou2013determining}, which is given by: 
\begin{equation}
\label{eq2}
RPS = \frac{1}{r-1} \sum _{i=1}^{r-1} \left( \sum _{j=1}^i (p_j - a_j)\right) ^2,
\end{equation}
where $r$ denotes the number of potential match outcomes (e.g., $r$ = 3
if there are three possible outcomes: home win, draw, and away win).
The RPS values always lie within the interval [0, 1], with a lower RPS indicating a better prediction.
In particular, an RPS value of 0 indicates a perfect prediction by a model, whereas a value of 1 represents a prediction that was completely incorrect.
%

\subsection{Approach 1: Deep Learning}
\label{sec:approach_dl}
\begin{figure}[]
    \centering
    \includegraphics[width=1\textwidth]{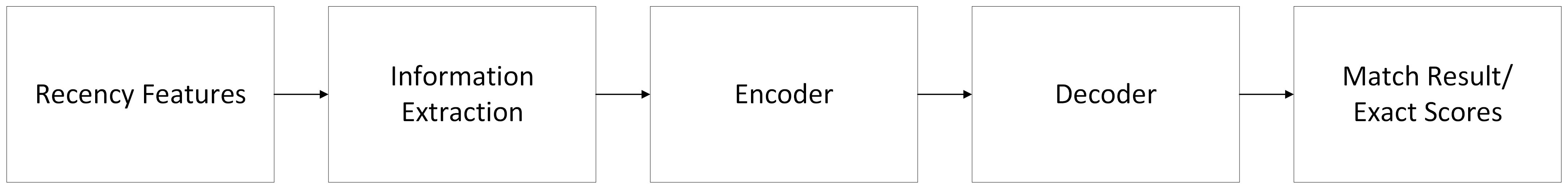}
    \caption{Refined concept of the deep learning approach.}
    \label{fig:dl_concept}
    \end{figure}

The deep learning approach utilizes time series-based features and a transformer-based model to predict the probabilities. Figure \ref{fig:dl_concept} depicts the refined concept of the deep learning approach.

\textbf{Step 1: Derive features for each match}. In 
 this step, we utilize the recency features extraction method proposed by \citep{berrar2019incorporating} in which, given a particular match of interest at time $t$, the $n$ previous matches at time $t-i$ of both the home and away teams are considered, where $i\in\{1,2,...,n\}$ and $n=5$. For each of the $t-i$ matches, the following features were derived: 
\begin{itemize}
    \item Team ID: a randomly assigned ID for the team
    \item Attacking strength: goals scored in match $t-i$.
    \item Defensive strength: goals conceded in match $t-i$.
    \item Strength of opposition: average goal difference of the opponent as of match $t-i$, calculated across its prior $n$ matches.
    \item Home advantage: a binary variable that takes a value of 1 if match $t-i$ was a home game and -1 if it was an away game.
\end{itemize}
The derived features can be represented as a matrix (Table \ref{tab:result_recency} shows the transpose of this matrix).

 \begin{table}[]
    \caption{Example of results from the recency feature extraction method. \citep{berrar2019incorporating}}
    \label{tab:result_recency}
    \begin{tabular}{clllll}
    \toprule

\multicolumn{1}{l}{}       &                        & \multicolumn{4}{c}{Recency} \\ \midrule
\multicolumn{1}{l}{}       & Feature group          & t-1   & t-2   & …    & t-n  \\ \midrule
\multirow{5}{*}{Home team} & Attacking strength     & 2     & 3     & ...  & 4    \\
                           & Defensive strength     & 0     & 0     & ...  & 2    \\
                           & Strength of opposition & 0.2   & 0     & ...  & 0.4  \\
                           & Home advantage         & -1    & 1     & ...  & -1   \\
                           & Team ID        & 101    & 101     & ...  & 101   \\                           
                           
                           \midrule
\multirow{5}{*}{Away team} & Attacking strength     & 0     & 0     & ...  & 2    \\
                           & Defensive strength     & 0     & 0     & ...  & 0    \\
                           & Strength of opposition & -0.4  & -0.4  & ...  & 0.6  \\
                           & Home advantage         & 1     & 1     & ...  & -1   \\
                           & Team ID        & 102    & 102     & ...  & 102   \\  
                           
    \botrule
    \end{tabular}
    \footnotetext{The value of the features are used for demonstrative purposes only and were retrieved from Table 2, \cite{berrar2019incorporating}.}
\end{table}

\textbf{Step 2: Information extraction}. Inspired by Timesblock in the TimesNet model \citep{wu2022timesnet}, we employ the inception block \citep{szegedy2015going} to extract additional features from the matrix from step 1, while retaining its shape. The inception block has traditionally been employed to allow subsequent layers to better capture information, and TimesNet has shown the effectiveness of applying the inception block to multi-dimensional time series. The matrix obtained in step 1 can be viewed as an 8-dimensional time series.

\textbf{Step 3 and 4: Encoding and Decoding}. In recent times, the Transformer Encoder (TE) \citep{vaswani2017attention} has become a popular method to embed soccer time series and sequential data into an informative vector \citep{yeung2023transformer,simpson2022seq2event}. This vector can then be decoded by a Multi-Layer Perception (MLP), to infer the target variable(s). In our case, we want to infer the probabilities of each of the possible match outcomes. Moreover, conventional Recurrent Neural Network models such as Long-Term Short-Term Memory (LSTM) \citep{hochreiter1997long} and Gated Recurrent Unit (GRU) \citep{chung2014empirical} could be used as the encoder. The performance of these encoders was compared with the TE in Section \ref{sec:prob_result}.

\subsection{Approach 2: Feature Selection and CatBoost}
\label{sec:approach_tree}
The feature selection and CatBoost approach makes use of features that were engineered in previous studies. It employs feature selection methods to identify the optimal feature set, which is then fed into the boosted decision tree model, CatBoost \citep{prokhorenkova2018catboost}.

\textbf{Step 1: Potential feature set formulation}. In the 2017 Soccer Prediction Challenge, the first- and second-placed participants: \cite{berrar2019incorporating} and \cite{hubavcek2019learning}, respectively, both employed gradient tree boosting models. Nevertheless, they utilized distinct sets of engineered features. Given this disparity, this study aims to investigate and identify the optimal feature set. We constructed a potential feature set by concatenating features from noteworthy methodologies employed in the 2017 challenge, as outlined by \cite{berrar2019incorporating} (1st and 5th place), \cite{hubavcek2019learning} (2nd place), and \cite{tsokos2019modeling} (4th place), as well as considering findings from other research focused on match result prediction \citep{baboota2019predictive}. The comprehensive compilation of potential features is tabulated in Table \ref{tb2}, totalling 205 distinct features.

\begin{sidewaystable}[]
\caption{Features considered for feature selection}
\label{tb2}
\begin{tabular}{lp{10cm}l}
\toprule
Features                         & Description                                                                                                                                 & Reference                                                                 \\
\midrule
GD                                 & Cumulative goal difference during the season                                                                                                &                                                                           \\
GS                                 & Cumulative goals scored during the season                                                                                                   &                                                                           \\
GC                                 & Cumulative goals conceded during the season                                                                                                 & \multirow{-3}{*}{N/A} 
\\
\midrule

elo                                & Elo ratings                                                                                                                                 & \url{https://footballdatabase.com/methodology.php} \\
\midrule

Streak                             & Performance in most recent (n=6) matches                                                                                                    &                                                                           \\
Weighted\_Streak                   & Performances in most recent (n=6) matches, weighted according to match recency                                                              &                                                                           \\
Form                               & Performance relative to opponents                                                                                                           & \multirow{-4}{*}{\cite{baboota2019predictive}}          \\
\midrule

attacking\_strength\_i              & Goals scored in match $t-i$, where $i$ $\in$ \{1,2,...9\}                                                                      &                                                                           \\
defensive\_strength\_i              & Goals conceded in match $t-i$, where $i$ $\in$ \{1,2,...9\}                                                                    &                                                                           \\
strength\_opposition\_i            & Home (away) team's opponent's average goal difference in their 9 most recent matches at time $t-i$, where i $\in$ \{1,2,...9\} &                                                                           \\
home\_advantage\_i                 & Whether the home team and away team played at home (1) or not (-1) in match at time $t-i$, where i $\in$ \{1,2,...9\}          &                                                                           \\
H\_Off\_Rating                     & Berrar rating home offensive rating                                                                            &                                                                           \\
H\_Def\_Rating                    & Berrar rating home defensive rating                                                                                             &                                                                           \\
A\_Off\_Rating                     & Berrar rating away offensive rating                                                                                          &                                                                           \\
A\_Def\_Rating                     & Berrar rating away defensive rating                                                                                &                                                                           \\
EG                                & Berrar rating predicted goals                                                                                                               & \multirow{-11}{*}{\cite{berrar2019incorporating}}         \\
\midrule

newly\_promoted                    & 1 if newly promoted, 0 otherwise                                                                                                            &                                                                           \\
newly\_demoted                     & 1 if newly demoted, 0 otherwise                                                                                                             &                                                                           \\
days\_since\_previous              & Number of days since the previous match                                                                                                     &                                                                           \\
Form2                              & Numbers of points gained in the last 3 matches, divided by 9                                                                                 &                                                                           \\
point\_tally                       & Number of points in the current season (up to but not including the current match)                                                                 &                                                                           \\
point\_per\_match                  & Average points scored per match in the current season (up to but not including the current match)                                                 &                                                                           \\
previous\_point\_tally             & Number of points in the previous season                                                                                                     &                                                                           \\
previous\_GS                       & Number of goals in the previous season                                                                                                      &                                                                           \\
previous\_GC                       & Number of goals conceded in the previous season                                                                                             &                                                                           \\
previous\_GD                       & Goal difference in the previous season                                                                                                      &                                                                           \\
days\_since\_first\_match\footnotemark[1]                  & Number of days since the first match in the league                                                                                          &                                                                           \\
quarter\footnotemark[1]                                    & Calendar year quarter                                                                                                                       & \multirow{-15}{*}{\cite{tsokos2019modeling}}          \\
\botrule

\end{tabular}
\footnotetext{All features are calculated for both the home team and the away team unless otherwise specified.}
\footnotetext[1]{Not based on the home or away team computed once only).}
\end{sidewaystable}

\setcounter{table}{1}
\begin{sidewaystable}[]
\caption{continued}
\begin{tabular}{lp{10cm}l}
\toprule
Features                         & Description                                                                                                                                 & Reference                                                                 \\
\midrule
L\_up\_i                           & Points difference from the first $i$ teams in the league table, where $i$ $\in$ \{1,2,…5\}                                     &                                                                           \\
L\_down\_i                         & Points difference from the bottom $i$ teams in the league table, where $i$ $\in$ \{1,2,…5\}                                     &                                                                           \\
Home\_venue\_win\_pct              & Home winning percentage in the current and last 2 seasons                                                                          &                                                                           \\
Away\_venue\_win\_pct              & Away winning percentage in the current and last 2 seasons                                                                         &                                                                           \\
win\_pct                           & Winning percentage in the current and last 2 seasons                                                                                        &                                                                           \\
Home\_venue\_draw\_pct             & Home draw percentage in the current and last 2 seasons                                                                             &                                                                           \\
Away\_venue\_draw\_pct             & Away draw percentage in the current and last 2 seasons                                                                            &                                                                           \\
draw\_pct                          & Draw percentage in the current and last 2 seasons                                                                                           &                                                                           \\
Home\_venue\_GS\_avg               & Average number of goals scored at home in the current and last 2 seasons                                                              &                                                                           \\
Away\_venue\_GS\_avg               & Average number of goals scored away in the current and last 2 seasons                                                             &                                                                           \\
GS\_avg                           & Average number of goals scored in the current and last 2 seasons                                                                            &                                                                           \\
Home\_venue\_GC\_avg              & Average number of goals conceded at home in the current and last 2 seasons                                                            &                                                                           \\
Away\_venue\_GC\_avg             & Average number of goals conceded away in the current and last 2 seasons                                                           &                                                                           \\
GC\_avg                            & Average number of goals conceded in the current and last 2 seasons                                                                          &                                                                           \\
home\_venue\_goal\_difference\_avg & Average goal difference at home in the current and last 2 seasons                                                                     &                                                                           \\
away\_venue\_goal\_difference\_avg & Average away goal difference in the current and last 2 seasons                                                                    &                                                                           \\
goal\_difference\_std              & Average goal difference in the current and last 2 seasons                                                                                   &                                                                           \\
Home\_venue\_GS\_std               & Standard deviation in the number of goals scored at home in the current and last 2 seasons                                            &                                                                           \\
Away\_venue\_GS\_std               & Standard deviation in the number of goals scored away in the current and last 2 seasons                                           &                                                                           \\
GS\_std                            & Standard deviation in the number of goals scored in the current and last 2 seasons                                                          &                                                                           \\
Home\_venue\_GC\_std               & Standard deviation in the number of goals conceded at home in the current and last 2 seasons                                          &                                                                           \\
Away\_venue\_GC\_std               & Standard deviation in the number of goals conceded away in the current and last 2 seasons                                         &                                                                           \\
GC\_std                            & Standard deviation in the number of goals conceded in the current and last 2 seasons                                                        & \multirow{-30}{*}{\cite{hubavcek2019learning}}                                                                          \\

\botrule

\end{tabular}
\footnotetext{All features are calculated for both the home team and the away team unless otherwise specified.}
\end{sidewaystable}

\setcounter{table}{1}
\begin{sidewaystable}[]
\caption{continued}
\begin{tabular}{lp{10cm}l}
\toprule
Features                         & Description                                                                                                                                 & Reference                                                                 \\
\midrule

home\_venue\_goal\_difference\_std & Standard deviation in the goal difference at home in current and last 2 seasons &                                                                           \\
away\_venue\_goal\_difference\_std & Standard deviation in the away goal difference in the current and last 2 seasons                                                  &                                                                           \\
goal\_difference\_std              & Standard deviation in the goal difference in the current and last 2 seasons                                                                 &                                                                           \\
win\_pct                           & Winning percentage in the last 5 matches                                                                                                    &                                                                           \\
draw\_pct                          & Draw percentage in the last 5 matches                                                                                                       &                                                                          \\
GS\_AVG                            & Average number of goals scored in the last 5 matches                                                                                        &                                                                           \\
GC\_AVG                            & Average number of goals conceded in the last 5 matches                                                                                      &                                                                           \\
GS\_STD                            & Standard deviation in the number of goals scored in the last 5 matches                                                                      &                                                                           \\
GC\_STD                            & Standard deviation in the number of home team goals conceded in the last 5 matches                                                                             &                                                                           \\
home\_venue\_goal\_scores\_avg\footnotemark[1]              & Average number of goals scored at home venues across the league in the last 2 seasons                                                      &                                                                           \\
away\_venue\_goal\_scores\_avg\footnotemark[1]              & Average number of goals scored at away venues across the league in the last 2 seasons                                                      &                                                                           \\
home\_venue\_goal\_scores\_std\footnotemark[1]              & Standard deviation in the number of goals per match scored at home venues across the league in the last 2 seasons                           &                                                                           \\
away\_venue\_goal\_scores\_std\footnotemark[1]              & Standard deviation in the number of goals scored per match at away venues across the league  in the last 2 seasons                         &                                                                           \\
home\_venue\_win\_pct\footnotemark[1]                       & Winning percentage across the league in the last 2 seasons                                                                                  &                                                                           \\
home\_venue\_draw\_pct\footnotemark[1]                      & Draw percentage across the league in the last 2 seasons                                                                                     &                                                                           \\
team\_cnt\footnotemark[1]                                  & Number of teams in the league in the previous season                                                                                        &                                                                           \\
gd\_std\footnotemark[1]                                    & Standard deviation in the goal difference across the league in the last 2 seasons                                                               &                                                                           \\
rnd\_cnt\footnotemark[1]                                   & Number of rounds in the league in the previous season                                                                                       &                                                                           \\
Round\footnotemark[1]                                      & The round in the current season                                                                                                                     &                                                                           \\
PageRank                           & PageRank computed based on the current and previous 2 seasons                                                                               &                                                                           \\
pi\_rating                         & Pi-rating computed based on the current and previous 4 seasons                                                                              & \multirow{-27}{*}{\cite{hubavcek2019learning}} \\
\botrule

\end{tabular}
\footnotetext{All features are calculated for both the home team and the away team unless otherwise specified.}
\footnotetext[1]{Not based on home or away team (computed once only).}
\end{sidewaystable}

\textbf{Step 2: Feature selection}. The process of feature selection is grounded in assessing, e.g., the correlation or information gain between input features and the target variable, the instance distances, and sometimes, the correlation among the different input features. Initially, four prevalent feature filtering techniques from WEKA were adopted, which take into account both the information content and correlation with the target variable. These include the Chi-square, Symmetrical Uncertainty, Correlation, and Information Gain attribute evaluation methods. Following consideration of the median ranking of the features across these filter methods, the 20 features most relevant were chosen. Subsequently, the ReliefF feature selection method \citep{kira1992practical,kononenko1994estimating} in the scikit-rebate library in Python \citep{Urbanowicz2017Benchmarking} was employed to factor in the distances between instances within the feature space. This resulted in an additional set of top 20 features. In the next phase, the elimination of duplicated features derived from the two aforementioned feature selection methods (filter and ReliefF) led to a compilation of up to 40 features. Ultimately, the Correlation Subset Feature Selection (CFS) method \citep{hall1988correlation} was utilized, which seeks to identify a feature set with the highest average correlation to the target feature, while simultaneously minimizing the average inter-feature correlation.


\textbf{Step 3: CatBoost}. While some recent studies \citep{berrar2019incorporating,hubavcek2019learning} have achieved strong performance with XGBoost \citep{chen2016xgboost} for soccer match result prediction, CatBoost is a more recent model that was introduced to overcome some of the limitations of XGBoost, especially in terms of the encoding of categorical features. CatBoost has similarities to XGBoost: both use gradient tree boosting, with regression or classification decision trees commonly used as weak learners. A unique aspect of CatBoost is its ordered target encoding. This method encodes the categorical features sequentially and without the use of the target feature, preventing any information leakage during training. Because of these advantages, in our approach, we decided on CatBoost as our chosen model. The CatBoost model was trained using the feature set mentioned in Step 2.

\section{Experiments and Results}
\label{sec:results}
In this section, we commence by elucidating the methodology for model training and validation, which is detailed in subsection \ref{sec:dataset_split}. Subsequently, we delve into the analysis and selection of models for Task 1 (exact score prediction) and Task 2 (probability prediction) in subsections \ref{sec:exact_result} and \ref{sec:prob_result}, respectively. Finally, in subsection \ref{sec:2023_result}, we deliberate upon the chosen model, and evaluate and discuss its performance relative to the top-placed performers in the 2023 Soccer Prediction Challenge.

\subsection{Model Training and Validation}
\label{sec:dataset_split}
In order to facilitate the training and validation of the models, the dataset was partitioned into three distinct training sets and three corresponding validation sets. For the three respective training datasets, five years' worth of data was utilized, encompassing seasons up to round $x-1$ of the 2018-19, 2019-20, and 2020-21 seasons. Conversely, the three respective validation sets were constructed using rounds $x$ and $x+1$ of the 2018-19, 2019-20, and 2020-21 seasons. The $x$ and $x+1$ correspond to the league rounds within the prediction set that spans the period from April 14 to April 26, 2023.

To ascertain the effectiveness of the proposed methodologies, i.e., 1) Deep Learning and 2) Feature Selection combined with CatBoost, the performance of each methodology was contrasted against baseline models, as well as ablated models, which investigate the effect of the exclusion or replacement of specific model components. For Task 1 (exact scores prediction), the baseline models comprised two simple statistical approaches, predicated on team and league-wide average goal scoring. Furthermore, models from previous research efforts were included, such as the Berrar ratings system \citep{berrar2019incorporating}, XGBoost applied to Berrar ratings \citep{berrar2019incorporating}, and CatBoost applied to pi-ratings \citep{razali2022football}.

For Task 2 (probabilities prediction), two simple baselines were used: team win/draw/loss (W/D/L) percentages, and a rule-based benchmark that always predicts a home team victory. Additionally, models from earlier literature were integrated, including the best-performing model from the 2017 Soccer Prediction Challenge --- Berrar ratings coupled with XGBoost \citep{berrar2019incorporating} --- alongside the best-performing model published in a study after the conclusion of the 2017 challenge, which applied a CatBoost model to pi-ratings features \citep{razali2022football}.

Furthermore, within the domain of Approach 1: Deep Learning, ablated models were explored, which, as mentioned, are models in which the effect of the exclusion or replacement of specific components is investigated. For instance, a model that excludes Information Extraction (TE+MLP), as well as models that substitute the encoder with LSTM or GRU.

All Models for Task 1 (exact scores prediction) and Task 2 (probabilities prediction) were trained to minimize Equations \ref{eq1} and \ref{eq2}, respectively. The hyperparameters for Approach 1 (Deep Learning) for Task 2 are listed in Supplementary Table \ref{tab:hyperparam}.

\subsection{Exact Scores Prediction Results}
\label{sec:exact_result}
In this subsection, the performance of exact score prediction using approaches 1 and 2 is compared with that of the baselines mentioned in Section \ref{sec:dataset_split}. The optimal feature set, chosen through approach 2, is presented in Supplementary Table \ref{tab:result_feature}.

 \begin{table}[]
    \caption{Exact scores prediction (Task 1) model results in terms of average model loss and standard deviation of model loss. The model that was preferred and used in the challenge is shown in \textbf{bold}.}
    \label{tab:result_exact}
    \begin{tabular}{lll}
    \toprule

Model                & Avg Loss & Sigma   \\
\midrule
\textbf{Berrar ratings}       & 1.0047   & 0.0434 \\
Team average   & 1.0206   & 0.0540 \\
XGBoost+Berrar ratings       & 1.0212   & 0.0381  \\
League average & 1.0346   & 0.0347  \\
CatBoost+selected feature set (Approach 2)   & 1.2162   & 0.0053  \\
CatBoost+pi-ratings   & 1.2356   & 0.0355  \\
TE+MLP (Approach 1)               & 1.5063   & 0.0317 
                           \\
    \botrule
    \end{tabular}
    \footnotetext{}
\end{table}

Through analyzing Table \ref{tab:result_exact}, it was evident that the Berrar ratings exhibited superior performance compared to the competing models. Notably, the team average statistical baseline followed, while the remaining models comprised primarily gradient tree boosting and deep learning models. This suggests that, despite the capability of gradient tree boosting and deep learning models in capturing complex relationships in data, these models proved unsuitable for accurately modeling exact match scores in soccer. In contrast, the Berrar ratings-based model, which incorporates team performance and domain knowledge related to matches, emerged as the more suitable choice for this task.

Given that Berrar ratings consistently outperformed the proposed approaches and other baselines, it was designated as the final model for Task 1 (exact score prediction). For further details on how the Berrar ratings are calculated, please refer to \cite{berrar2019incorporating}.

\subsection{Probabilities Prediction Results}
\label{sec:prob_result}
The performance of approaches 1 and 2 in probabilities prediction was then compared to the baseline models and ablated models previously described in subsection \ref{sec:dataset_split}. The optimal feature set selected in Approach 2 is listed in Supplementary Table \ref{tab:result_feature}.

\label{sec:prob_result}
 \begin{table}[]
    \caption{Probabilities prediction (Task 2) model results in terms of average model loss and standard deviation of model loss. The model that was preferred and used in the challenge is shown in \textbf{bold}.}
    \label{tab:result_prob}
    \begin{tabular}{llllll}
    \toprule

Model                           & Avg Loss & Sigma   \\ 
\midrule
CatBoost+pi-ratings              & 0.2085   & 0.0083        \\
\textbf{Inception+TE+MLP (Approach 1)}       & 0.2098   & 0.0051        \\
LSTM+MLP                        & 0.2105   & 0.0050        \\
TE+MLP                          & 0.2111   & 0.0062        \\
GRU+MLP                         & 0.2116   & 0.0052        \\
XGBoost+Berrar ratings           & 0.2141   & 0.0046        \\
W/D/L percentage & 0.2303   & 0.0015        \\
CatBoost+selected feature set (Approach 2)      & 0.2416   & 0.0028        \\
Home win                        & 0.4450   & 0.0031         \\
    \botrule
    \end{tabular}
    \footnotetext{}
\end{table}

By examining Table \ref{tab:result_prob}, it becomes evident that among the distinct models, the CatBoost+pi-ratings model demonstrated the highest performance, with the lowest average loss. Following closely was the Inception+TE+MLP model (Approach 1), outperforming the XGBoost+Berrar ratings and other remaining models. Despite the superior performance of the CatBoost+pi-ratings model, it is noteworthy that its loss on the 2018-19 season validation set was 0.1991, lower than the loss of 0.2072 achieved by Approach 1 by a margin of 0.0083. Meanwhile, the performance observed across the other two validation sets had a smaller margin of $\pm$ 0.018. In particular, while the average loss of the Inception+TE+MLP model (Approach 1) was not much greater than that of CatBoost+pi-ratings, the standard deviation of the model loss (0.0051) of the former was markedly lower than that of the latter (0.0083). Considering the possibility of the 2018-19 season being an exceptional case, Approach 1 was deemed the preferred final model for the challenge.

Upon further scrutiny of the ablated version of Approach 1, the model devoid of the inception block (TE+MLP) displayed weaker performance. Moreover, in comparing the encoder LSTM, TE, and GRU, the LSTM exhibited the most favorable performance. However, due to the considerable training time required for hyperparameter grid searching in LSTM, TE was selected as a more practical alternative given the limited time available to meet the competition deadline.

Lastly, the performance of the CatBoost+Selected feature set model fell short of the performance of the CatBoost+pi-ratings model. Consequently, the prospect of augmenting the model with additional engineered features seemed less promising. Instead, focusing on enhancing pi-ratings appeared a more viable strategy for further work. Nevertheless, the incorporation of additional engineered features in Approach 1 is also an avenue for potential further research.

\subsection{2023 Soccer Prediction Challenge Comparative Evaluation}
\label{sec:2023_result}
The final model was trained using 5 years of data until April 14, whereas the provided training set only provided data until April 4, 2023. To address this gap, as previously mentioned, data from April 4, 2023, to April 14, 2023, was manually appended to the training set.

Our performance in the 2023 Soccer Prediction Challenge in the two required challenge tasks, compared to the top-performing team, is summarized in Tables \ref{tab:spc_result_task1} and \ref{tab:spc_result_task2}. In Task 1, the Berrar ratings \citep{berrar2019incorporating} were surpassed by the first-place team by a margin of 11.91\%. As for Task 2, our Approach 1 involving deep learning was outperformed by a bookmaker consensus type model by 6.42\%. Given that the 2023 Soccer Prediction Challenge permitted the utilization of alternative features and training instances beyond the provided dataset, the integration of additional features stands as a potential avenue for enhancing model performance. This is particularly relevant to Approach 1 for Task 2, where alternative features could be seamlessly incorporated.

\begin{table}[]
    \caption{Task 1 result compared to the top-performing approach}
    \label{tab:spc_result_task1}
    \begin{tabular}{ll}
    \toprule

Team               & RMSE  \\
\midrule
TeamNateWeller      & 1.6235  \\
Berrar ratings (Ours)      & 1.8169  \\
    \botrule
    \end{tabular}
    \footnotetext{}
\end{table}

\begin{table}[]
    \caption{Task 2 result compared to the top-performing approach}
    \label{tab:spc_result_task2}
    \begin{tabular}{ll}
    \toprule

Team               & RMSE  \\
\midrule
Bookmakers      & 0.2063  \\
Approach 1 (Ours)      & 0.2195  \\
    \botrule
    \end{tabular}
    \footnotetext{}
\end{table}


Upon scrutinizing the validation and results from the 2023 soccer prediction challenge, it is apparent that, in terms of exact score prediction, machine learning models presently lag behind rule-based counterparts such as team ratings. Furthermore, the consideration of alternative features, encompassing elements like expert opinions, team formations, and player details --- elements potentially encapsulated in betting odds --- has the potential to bolster the efficacy of deep learning models in match outcome probability prediction.

\section{Related Work}
\label{sec:related}
In this section, we review related research on deep learning for match results prediction in soccer and also studies that have used the Open International Soccer Database \citep{dubitzky2019open} to build their models.

Despite its successful application in a number of application domains, there are still relatively few studies that have applied deep learning models for soccer match result prediction.
\cite{danisik2018football} used an LSTM model to predict soccer match outcomes, comparing classification, numeric prediction, and dense approaches, and also with baselines based on the average random guess, bookmaker odds-derived predictions, and home win (the majority class).
Data from the English Premier League was used, and player-level data was obtained from the FIFA video game for the classification and numeric prediction approaches.
The average accuracy obtained with the LSTM regression model was 52.5\%.
%
\cite{jain2021soccer} used Recurrent Neural Networks and LSTM networks for soccer match result prediction.
Using English Premier League data, the authors manually engineered several relevant features, e.g., winning and losing streaks, points, and goal differences.
Their reported accuracy was 80.75\%, however, it should be noted that this was for 2-class, not a 3-class prediction.
\cite{rahman2020deep} used deep neural networks and artificial neural networks for soccer match result prediction using primarily rankings and results, achieving 63.3\% accuracy when predicting 2018 FIFA World Cup matches.
Recently, \cite{malamatinos2022predicting} used k-Nearest-Neighbors, LogitBoost, Support Vector Machine, Random Forest, and CatBoost, along with Convolutional Neural Networks and Transfer Learning with tabular data that was encoded and converted to image models, to predict the results of Greek Super League matches. 
The best-performing model, CatBoost, which was found to outperform the Convolutional Neural Network, was also applied to predict English Premier and Dutch Eredivisie league matches.
Also recently, \cite{joseph2022time} used statistical time series approaches to
predict English Premier League outcomes and compared their performance with LSTM and Bayesian methods.

In more recent times, driven by the popularity of the Transformer model \citep{vaswani2017attention}, attention mechanisms have gained prominence in modeling soccer-related data. Notably, attention mechanisms have been harnessed in various ways for soccer data analysis. For instance, \cite{zhang2022sports} introduced an attention-based LSTM network to predict soccer match result probabilities. Another notable contribution comes from \cite{simpson2022seq2event}, who proposed the Seq2event model, based on the transformer architecture, for modeling sequential soccer data. This work was subsequently extended and refined by \cite{yeung2023transformer}. Given this backdrop, this study aimed to investigate the effectiveness of attention mechanisms in the context of soccer match results prediction. Furthermore, we intended to delve into investigating the potential of state-of-the-art time series models, leveraging the inherent sequential nature of soccer match results data.

The standard evaluation metric used in the 2017 Soccer Prediction Challenge, and subsequent studies that have used the Open International Soccer Database, has been the Ranked Probability Score (RPS) \citep{epstein1969scoring,constantinou2012solving}.
However, other evaluation metrics such as cross-entropy \citep{hubavcek2022forty} and accuracy have also been used.
Since the RPS is sensitive to distance \cite{constantinou2012solving}, where the ordered nature of win/draw/loss has been considered, the RPS is a preferable metric in evaluating match outcome prediction model performance.
However, subsequently, \cite{wheatcroft2021evaluating} suggested that the ignorance score may be a more appropriate metric for match outcome prediction model evaluation.

Among the studies from 2017 Soccer Prediction Challenge participants, \cite{tsokos2019modeling} used a Bradley–Terry model, Poisson log-linear hierarchical model, and an integrated nested Laplace approximation, achieving an RPS of 0.2087 and accuracy of 0.5388.
\cite{hubavcek2019learning} used relational- and feature-based methods, with pi-ratings \citep{constantinou2013determining} and PageRank ratings \citep{page1998pagerank} computed for each of the teams in each match.
XGBoost \citep{chen2016xgboost} was employed as the feature-based method and was used for both classification and regression, and boosted relational dependency networks (RDN-Boost) \citep{natarajan2012gradient} were used as the relational method.
Classification with XGBoost, achieving RPS and accuracy of 0.2063 and 0.5243, respectively, performed best on the validation set and the challenge test set.
%
\cite{constantinou2019dolores} proposed a Hybrid Bayesian Network, using dynamic ratings based on the pi-rating system developed in previous work \citep{constantinou2013determining} but that also incorporated a team form factor to identify continued over- or under-performance.
In the modified pi-rating calculation, the (win, draw, loss) match outcome was emphasized to a greater extent than the goal margin in order to dampen the effect of large goal margins.
The Hybrid Bayesian Network was applied to four rating features --- two each for the home and away teams. 
The model was able to make accurate predictions for a match between two teams even when the prediction was based on historical match data that involved neither of the two teams, with the model achieving accuracy of 0.5146 and an RPS of 0.2083 on the challenge test data set. 
\cite{berrar2019incorporating} created two types of feature sets: recency and rating features.
Recency features consisted of the averages of features over the previous nine matches, based on four feature groups: attacking strength, defensive strength, home advantage, and opposition strength. 
%
%
XGBoost and k-Nearest-Neighbors were applied to each of the two feature sets, with both models performing better on the rating features than the recency features.
XGBoost, when applied to the rating features, provided the best performance, although this result of 0.5194 accuracy and 0.2054 RPS was obtained after the competition had concluded. 
The k-Nearest-Neighbors model applied to the rating features, which achieved accuracy of 0.5049 and an RPS of 0.2149 on the competition test set, was the best result achieved during the competition.

The top-performing participants in the 2017 Soccer Prediction Challenge commonly applied machine learning to rating features.
At least in the absence of match-related features on in-game events, ratings, therefore, seem to be an effective means of condensing a large amount of historical match information into a concise set of model features.
It was also evident that gradient-boosted tree models such as XGBoost exhibited strong performance in the competition.
%

Subsequent to the 2017 Soccer Prediction Challenge, other researchers have made use of the Open International Soccer Database.
\cite{robberechts2019forecasting} compared the performance of result-based Elo ratings and goal-based offensive-defensive models in predicting match results in FIFA World Cup and Open International Soccer database matches.
The ELO ordered logit achieved an RPS of 0.2035 and an accuracy of 0.5146, while the
ELO plus offensive-defensive model ordered logit obtained RPS and accuracy of 0.2045, and 0.5146, respectively. However, bookmaker odds-obtained predictions achieved slightly better performance, with a lower RPS of 0.2020 and slightly higher accuracy of 0.5194.
\cite{razali2022football} also used the Open International Soccer Database, comparing the performance of gradient-boosted tree models such as XGBoost, LightGBM, and CatBoost on goal- and result-based Elo ratings and pi-ratings.
The authors found that CatBoost applied to the pi-ratings features yielded the best performance (RPS = 0.1925), which was better than the results achieved by the 2017 Soccer Prediction Challenge participants.
\cite{razali2022deep} used a deep learning-based approach by applying TabNet, an interpretable canonical deep tabular data learning architecture, to pi-ratings and achieved a slightly higher RPS of 0.1956, which was still better than the 2017 Soccer Prediction Challenge participants.
\cite{hubavcek2022forty} compared several statistical Bivariate and Double Poisson and Weibull distributions-based statistical models and ranking systems, specifically, Elo ratings, Steph ratings, Gaussian-OD ratings, as well as the soccer-specific rating systems of Berrar ratings and pi-ratings.
Through their experiments using the Open International Soccer database --- with matches before July 2010 forming a validation set that was used for hyperparameter tuning, and matches after this date forming a test set of 91,155 matches --- the authors found that Berrar ratings provided the lowest RPS (0.2101) among the different statistical models and rating systems.
However, the other models also provided performance very close to that of Berrar ratings, suggesting the existence of some limits to match prediction performance.
Since \cite{hubavcek2022forty} did not use any boosting or deep learning methods, we can tentatively conclude that --- at least on datasets such as the Open International Soccer Dataset that do not contain features derived from match events other than the goals scored --- gradient-boosted tree models and/or deep learning models can outperform the rating systems themselves and statistical model.

In this study, we devised two distinct approaches. Firstly, in the context of the gradient-boosted tree model, it is noteworthy that several methods have demonstrated superior performance by leveraging unique features. Consequently, our primary emphasis within this approach lies in pinpointing the optimal feature set through the utilization of feature selection algorithms. Secondly, taking into account the remarkable achievements of deep learning models in recent years, our objective was to probe the potential of these models in the realm of predicting soccer match outcomes. This endeavor involved an exploration of the predictive prowess of deep learning methodologies.

\section{Conclusion}

\label{sec:conclusion}
In this study, our objective was to assess the performance of a deep learning model and determine the optimal feature set for a gradient-boosted tree model in predicting soccer match results in terms of win/draw/loss (W/D/L) probabilities as well as exact scores. To achieve this aim, we introduced a deep learning-based model and leveraged features from prior studies, coupled with feature selection algorithms, to identify the most effective feature set. Our models were trained, validated, and tested using data from the 2023 Soccer Prediction Challenge. The results revealed that, in terms of W/D/L probabilities, the deep learning model was outperformed by betting odds consensus model predictions by a margin of 6\%. Moreover, it was noteworthy that pi-ratings still retained their status as the most suitable features for using with gradient-boosted tree models. When it comes to predicting exact scores, the Berrar ratings and simple statistical baseline models exhibited superior performance compared to both the deep learning and gradient-boosted tree models.

Future endeavors could focus on enhancing model interpretability. Given that both deep learning and boosted decision tree models fall under the category of black-box models, efforts to enhance interpretability would greatly benefit coaches and analysts in identifying the pivotal features for achieving victory. Furthermore, consideration could be given to alternative features, such as betting odds that encapsulate expert opinions, team compositions, player characteristics, and more. Despite this, we anticipate that our study will underscore the efficacy of employing deep learning methodologies in predicting soccer match outcomes, thereby inspiring forthcoming research in this domain.

\backmatter

\bmhead{Acknowledgments}
This work was financially supported by JST SPRING (Grant Number JPMJSP2125). 
Calvin Yeung would like to take this opportunity to thank the “Interdisciplinary Frontier Next-Generation Researcher Program of the Tokai Higher Education and Research System.”

\section*{Declarations}
\begin{itemize}
\item Funding: This work was financially supported by JST SPRING, Grant Number JPMJSP2125.
\item Competing interests: The authors have no competing interests to declare that are relevant to the content of this article.
\item Ethics approval: Not applicable
\item Consent to participate: Not applicable
\item Consent for publication: Not applicable
\item Availability of data and materials: The data of the research was collected and provided by the 2023 Soccer Prediction Challenge 

\item Code availability: The code for the model is available at \url{https://github.com/calvinyeungck/Soccer-Prediction-Challenge-2023}
\item Authors' contributions: All authors contributed to the study conception and design. Data preparation, modeling, and analysis were performed by all authors. The first draft of the manuscript was written by all authors and all authors commented on previous versions of the manuscript. All authors read and approved the final manuscript.
\end{itemize}

\begin{appendices}
\setcounter{table}{6}
\setcounter{figure}{0}
\section{Model Hyperparameters}

\begin{table}[h]
\caption{Optimal Hyperparmeters for Task 2 Approach 1}
\label{tab:hyperparam}
\begin{tabular}{lll}
\toprule
Hyperparameter          & Grid-searched Value        & Optimal Value \\
\midrule
team\_id\_embedding\_dim & 1,2,4,8,16                 & 1             \\
TE\_dim\_feedforward     & 1,8,64,512,2048,4096       & 1             \\
TE\_dropout              & 0,0.1,0.2,0.3                  & 0             \\
MLP\_num\_layer          & 1,2,3,4,5,6,7,8,9,10,11,12 & 10            \\
MLP\_dropout             & 0,0.1,0.2,0.3              & 0.2      \\
\botrule
\end{tabular}
\end{table}

\section{Approach 2 Selected Feature Set}
\begin{table}[h]
    \caption{Optimal feature set from Approach 2}
    \label{tab:result_feature}
    \begin{tabular}{ll}
    \toprule

Target Feature                & Optimal Feature Set  \\
\midrule
Home team goals      & EG\_HT, GS\_avg\_HT   \\
Away team goals      & Home\_venue\_GS\_avg\_AT, GC\_avg\_HT, pi\_rating\_AT, GC\_AVG\_HT, previous\_GD\_AT   \\
W/D/L probability     & EG\_HT, EG\_AT, point\_per\_match\_HT, win\_pct\_AT, pi\_rating\_HT, pi\_rating\_AT\\

    \botrule
    \end{tabular}
    \footnotetext{An explanation of the features is available in Table \ref{tb2}. HT or AT denotes the features that are calculated based on Home Team or Away Team, respectively.}
\end{table}




\end{appendices}

\bibliography{sn-bibliography}

\end{document}